\documentclass[acmsmall,screen]{acmart}

\usepackage{CJKutf8}
\usepackage{bm}
\AtBeginDocument{%
  \providecommand\BibTeX{{%
    \normalfont B\kern-0.5em{\scshape i\kern-0.25em b}\kern-0.8em\TeX}}}

\begin{document}

\title{Chinese Spelling Error Detection Using a Fusion Lattice LSTM}


\author{Hao Wang}
\email{wanghaomails@gmail.com}
\author{Bin Wang}
\author{JianYong Duan}
\affiliation{%
  \institution{North China University of Technology}
  \streetaddress{ShiJingShan district JinYuanZhuang No.5}
  \city{BeiJing}
  \country{China}}

\author{JiaJun Zhang}
\affiliation{%
 \institution{National Laboratory of Pattern Recognition, Institute of Automation,
Chinese Academy of Sciences, University of Chinese Academy of Sciences}
 \streetaddress{7-th Floor, Intelligence Building,
No. 95, Zhongguancun East Road, Haidian District}
 \city{Beijing}
 \state{Beijing}
 \country{China}}






\begin{abstract}
Spelling error detection serves as a crucial preprocessing in many natural language processing applications. Due to the characteristics of Chinese Language, Chinese spelling error detection is more challenging than error detection in English. Existing methods are mainly under a pipeline framework, which artificially divides error detection process into steps. Thus, these methods bring error propagation  and cannot always work well due to the complexity of the language environment. Besides existing methods only adopt character or word information, and ignore the positive  effect of fusing character, word and pinyin\footnote{pinyin is a system of writing Chinese in Latin characters which was introduced in China in the late 1950s. The term in Chinese means literally ‘spell-sound’} information together. We propose an
FL-LSTM-CRF model, which is an extension of the LSTM-CRF
with word lattices and character-pinyin-fusion inputs. Our model  takes  advantage of the end-to-end framework to detect errors as a whole process, and dynamically integrates character, word and pinyin information. Experiments on the SIGHAN data show that our FL-LSTM-CRF  outperforms existing methods with similar external resources consistently, and  confirm the feasibility of adopting the
end-to-end framework and the availability of integrating of
character, word and pinyin information.
\end{abstract}

\ccsdesc[500]{Computer systems organization~Embedded systems}
\ccsdesc[300]{Computer systems organization~Redundancy}
\ccsdesc{Computer systems organization~Robotics}
\ccsdesc[100]{Networks~Network reliability}

\keywords{datasets, neural networks, gaze detection, text tagging}

\maketitle

\begin{CJK*}{UTF8}{gbsn}
\section{Introduction}
\label{sec:introduction}

Spelling error detection is a common task in every written language, which aims to detect human errors \cite{Wu2013ChineseSC}, and 
is a vital prerequisite step for many natural language processing  applications, 
such as search engine \cite{Martins2004SpellingCF,Gao2010ALS},
automatic essay scoring system  \cite{Burstein1999AutomatedES,Lonsdale2003AutomatedRO} and so on.

Chinese spelling check is very different from English due to several distinct characteristics of Chinese language.

1.	There are more than 100,000 Chinese characters, and about 3,500 are frequently used in daily life. Many Chinese characters have similar shapes and/or similar pronunciations.

2.	There are no delimiters between words. Word is not a natural concept for Chinese Language, and we have to use a word segmentation tool to obtain words.

3.	In English, every single word is directly typed by the keyboard. The most frequent errors refer to misspelled words which cannot be found in any dictionary \cite{Yu2014ChineseSE,hsieh2015correcting}, called “non-word errors”. However, Chinese characters cannot be typed directly, thus we have to use the  input method, which just attempts to map a sequence of Latin letters to the character stored in a dictionary of the computer. All  Chinese characters can be found in a dictionary and Chinese errors are “real-word errors”.

Because of the above characteristics of language, Chinese
spelling check becomes a more challenging task. Based on the observation that words with spelling errors can often be divided into single-character words using word segmentation tools.  For example, “交洁” (交{ } is a spelling error) will be divided into two single-character words “交” and “洁”. Existing
methods  often  detect errors under  the pipeline framework. They first divide the input sentence using word segmentation tools, and then treat all single-character words as error candidates. Replace each of these candidates with its corresponding words in a confusion set \footnote{A confusion set refers to a set that contains a character and its corresponding confusing characters. Characters and their corresponding characters often have similar shape or pronunciation. Take “交” as an example. The confusing characters are 皎,叫,焦{ } and etc.}. Finally, a language model is used to score these sentences before and after the replacement. If the score of the sentence after the replacement is higher, the character before the replacement is identified as the error word, and the word from the confusion set is the correction. 

\begin{figure}[h]
	\centering
	\includegraphics[width=0.6\linewidth]{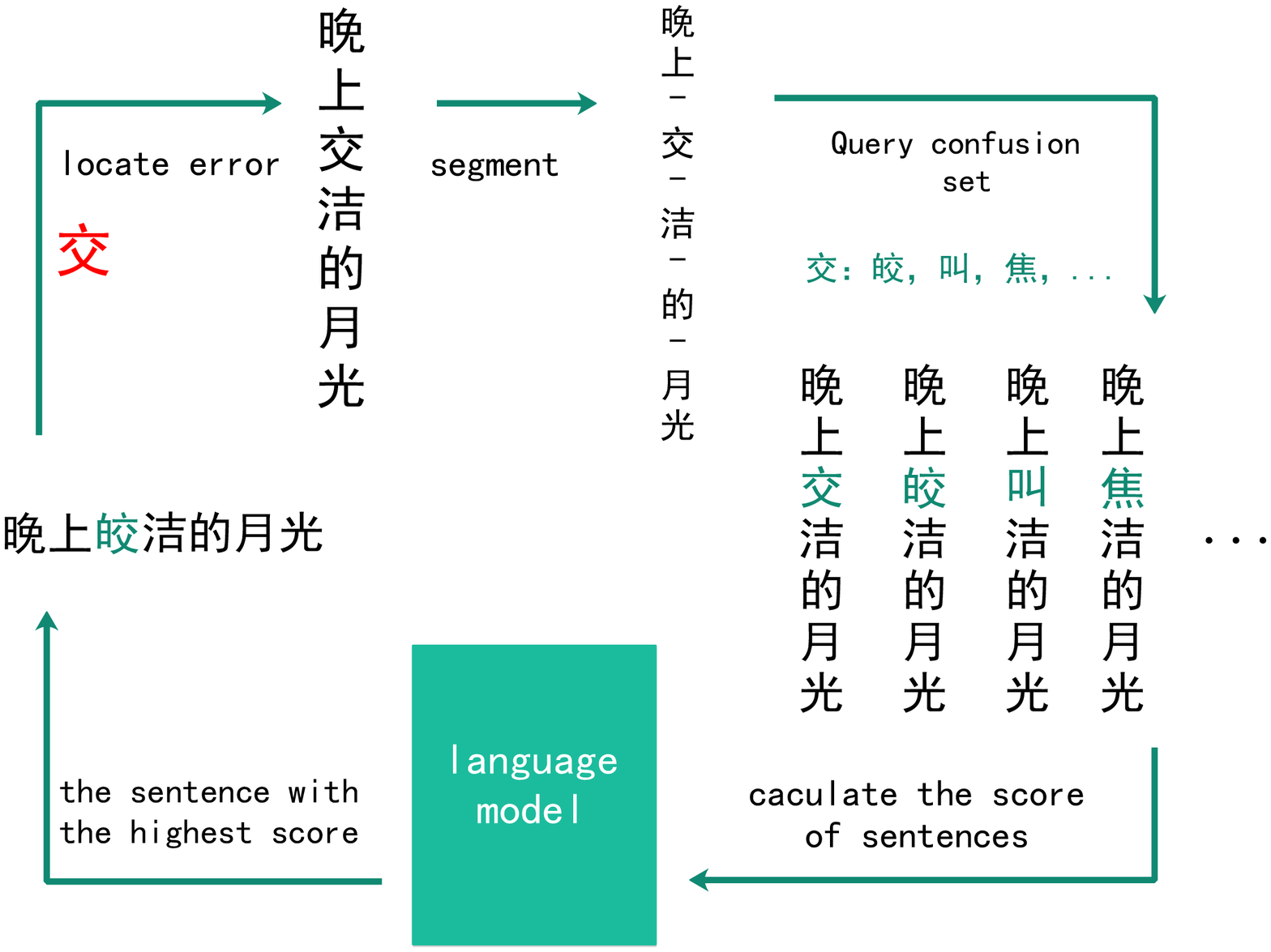}
	\caption{Detecting spelling errors in  pipeline}
	\label{fig:pipeline} 
\end{figure}

Figure \ref{fig:pipeline} shows the process of existing  models to detect spelling errors of the  sentence “晚上交洁的月光 (the bright moonlight at night)”. Firstly, a word segmentation tool performs on the input sentence to obtain the segmented sentence “晚上/交/洁/的/月光”, in which “交”, “洁”, “的” are single-character words. Taking “交” as an example, query the confusion set  to obtain its corresponding confusion items “皎,叫,焦,...”, replace “交” with each of confusion items and send the sentences before and after the replacement to a language model for scoring. Searching the sentence with the highest score from all  sentences after and before the replacement, we locate the spelling error and correction. For example, if the sentence “晚上皎洁的月光” with the replaced item “皎” obtains the highest score, the original “交” is judged as a spelling error and “皎” is the correction.

Existing  methods mainly have the following shortcomings. First,
existing methods are based on a pipeline architecture, and artificially divide the error detection process into two parts: querying error candidates and locating the errors. This two-stage detection mode is easy to cause error propagation, and it cannot always work well due to the complexity of the language environment.
Second, existing methods mainly adopt the character or word information separately, and ignore pinyin information. 





\begin{figure*}[t]
\centering
	\includegraphics[width=0.89\linewidth]{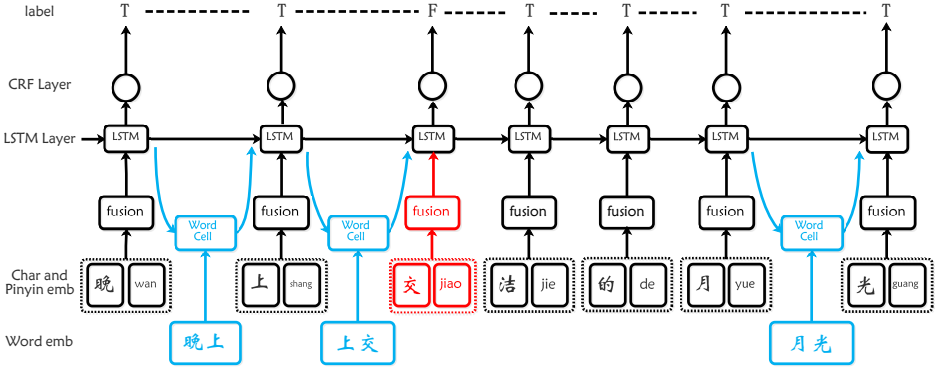}
	\caption{ The framework of FL-LSTM-CRF model. The blue part represents word lattice. The fusion cell represents the input gates of character and pinyin information. The red part represents the misspelled character.}
	\label{fig:model}
\end{figure*}

To solve the above issues, We propose an
FL-LSTM-CRF model for error detection, which is an extension of the LSTM-CRF
with word lattices and character-pinyin-fusion inputs. As shown in Figure \ref{fig:model}, for each input character sequence, we calculate the fusion input by dynamic mixing of character and pinyin information, and find all subsequences that are matched words in the word vocabulary. Finally, the fusion input will be feed to the model as an new input, and these subsequences (called word lattices) will link different LSTM cells as information shortcuts. Our main contributions are as follows:

1. We propose a model under the end-to-end framework to detect spelling errors, which treats the error detection as a whole process, and automatically extracts useful information avoiding artificial intervention.

2. We introduce a novel way to dynamically integrate character,  word  and  pinyin  information,  which  overcomes  the shortcomings of only using the single information and improves the flexibility of extracting information.

3.  Experiments  on  SIGHAN  data show our FL-LSTM-CRF model outperforms all existing models with similar external resources. The end-to-end framework is an excellent choice for error detection.
Integrating character, word and pinyin information dynamically will effectively improve the performance.







\section{Related Work}
\label{sec:related-work}

Currently, most Chinese spelling error detection can be roughly divided into three categories according to the information in use.

In the first category, character information is used.   \cite{chang1995new} 
proposed a method which replaced each character in a sentence by 
characters from a confusion set, and calculated the best scored 
sentence within the original sentence and all replaced sentences 
by a bi-gram language model. The spelling errors 
were detected by comparing the best scored sentence with the original 
one. \cite{han2013maximum} detected each character spelling 
error in a sentence using a maximum entropy model trained on a large corpus. In \cite{yeh2013chinese}, they generated a dictionary containing similar pronunciation and form for every Chinese character, and then proposed a system based on this dictionary to detect spelling errors.

In the second category, word information is used. \cite{liu2013hybrid} used a word segmentation based language model and 
a statistical machine translation model to search for errors, and then verifying the 
errors by reranking the correction candidates. In their model, the translation probability indicated how likely a typo was translated into its correct word.
\cite{chen2013study} implemented a new forward and backward word segmentation tool without automatically merging algorithm to search for errors. Confusion sets and various language models (n-gram, topic modeling and document topic modeling) were used to verify errors. 
\cite{chiu2013chinese} segmented sentences to generate the sequences of two or more singleton words. These two or more singleton words were treated as errors, and verified by a Chinese dictionary and web-based ngrams. 
\cite{hsieh2013introduction} developed two segmentation systems with different dictionaries to search for errors. A simple
maximizing tri-gram frequency model based
on Google 1T tri-gram was designed to verify errors
and select the correct answers.  
In \cite{yang2013sinica}, they proposed a high-confidence pattern matcher to search for errors after segmentation. These errors were verified and corrected by the rest component of the system.

In the third category, character and word information is used. In \cite{he2013description}, three methods were proposed to detect character-level 
errors , word-level errors, and context-level errors, respectively. 



Our work is inspired by  
\cite{zhang_chinese_2018,yang_subword_2018}. They utilized lattice models to recognize Chinese name entity and segment Chinese sentences. Our work is an extension of the above work in the field of Chinese error detection, which introduces a dynamic fusion mechanism to accommodate characteristics of Chinese language.


\section{Models}
\label{sec:models}

We treat Chinese spelling error detection as a sequence tagging problem and take the state-of-the-art LSTM-CRF without CNN layer as a baseline model \cite{ma_end--end_2016}. Formally, given an input sentence s with m Chinese characters $s = (c_1,...,c_t,..., c_m$), our model is to assign each character with a label $l_t$, where $l_t  \in \{T,F\}$. The labels T and F represent the true character and error character, respectively \cite{wang_hybrid_2018}.    Figure \ref{fig:model} shows the framework of our spelling error detection model on an input “晚上交洁的月光 (the bright moonlight at night)” , where the black part represents the baseline LSTM-CRF model, the blue part  and the green part represent  the word lattice and the pinyin lattice, respectively.  The spelling error “交” of the input is marked in red and labeled with F.

\subsection{Embedding Layer and Fusion Inputs}
\subsubsection{Embedding Layer} As shown in Figure \ref{fig:model}, our model includes three embeddings, namely character embedding,  pinyin embedding for characters and word embedding. For each  character $c_e$, the corresponding character representation and pinyin representation are calculated as follows: 
\begin{equation}
\label{eq:char_py_emb}
\begin{split}
\bm{x}_{e}^{c}&=\bm{e}^{c}\left(c_{e}\right)\\
\bm{x}_{e}^{p}&=\bm{e}^{p}\left(c_{e}\right)\\
\end{split}
\end{equation}
where $e^c$ and $e^p$ denote the character and pinyin embedding lookup table, respectively.

For each character subsequence $(c_b...c_e)$ recorded as $c_{b:e}$. The corresponding word lattice representation is calculated as follows,
\begin{equation}
\label{eq:word_emb}
\begin{split}
\bm{x}_{b,e}^{w}&=\bm{e}^{w}\left(c_{b:e}\right)
\end{split}
\end{equation}
where $x^w_{b,e}$ denotes the word representation. $e^w$ denotes the word  embedding lookup-table. b denotes the start index  of the character subsequence in a whole sequence, e the end index. 

\begin{figure}[h]
	\centering
	\includegraphics[width=0.5\linewidth]{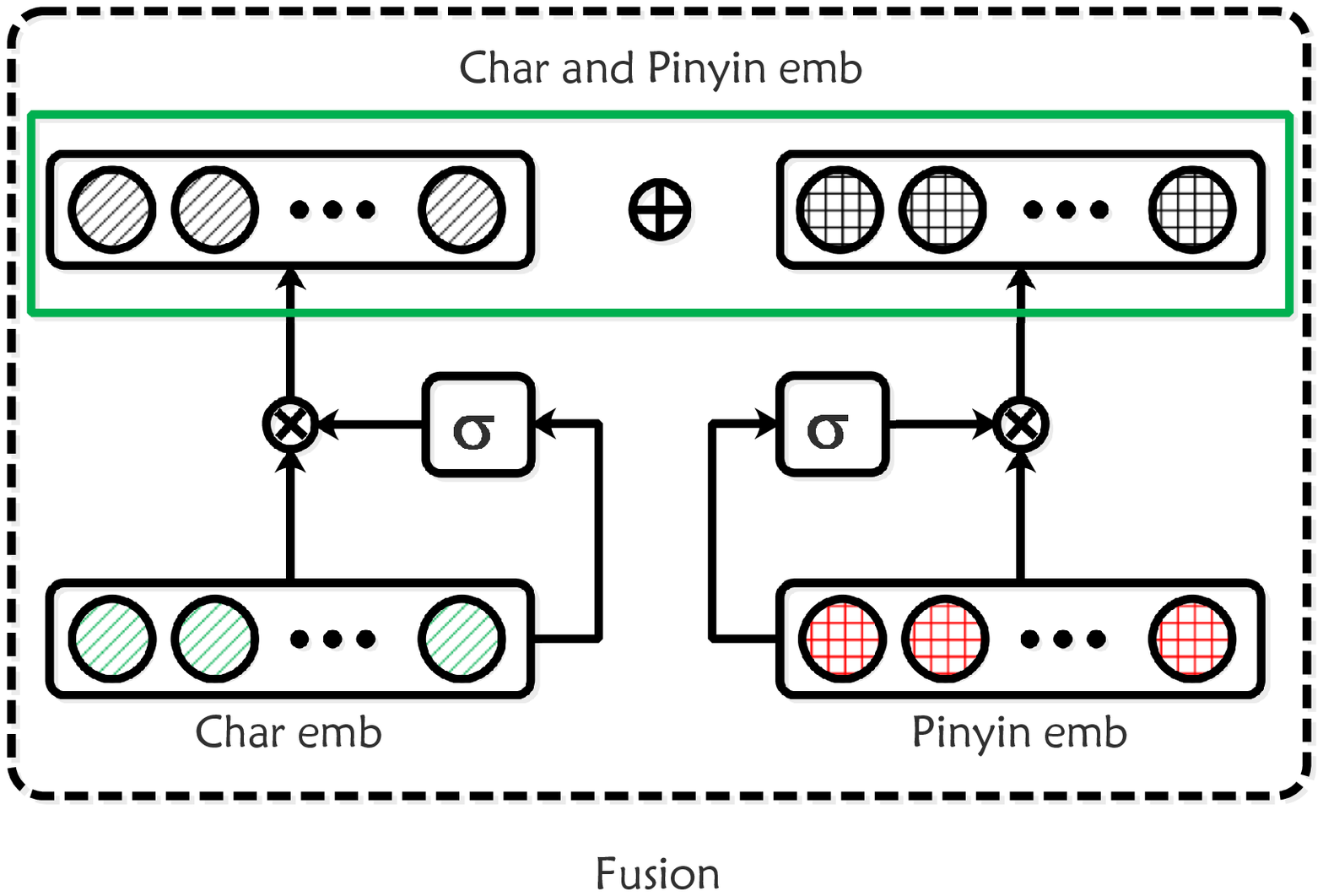}
	\caption{The structure of fusion cell}
	\label{fig:fusion-inputs}
\end{figure}

\subsubsection{Fusion Inputs} 
Character and pinyin describe the sentence from text and pronunciation aspects, respectively. In order to fuse both information, we propose the sample-specific gate, as shown in Figure \ref{fig:fusion-inputs}, to  assign different weights to character and pinyin dynamically. The weight parameters are calculated as follows:  
\begin{equation}
\label{eq:input_gate}
\begin{split}
\bm{g}^c_{e}=\sigma({W^c_{g}}^T\bm{x}^c_e+b^c_{g})\\
\bm{g}^p_{e}=\sigma({W^p_{g}}^T\bm{x}^p_e+b^p_{g})
\end{split}
\end{equation}
where ${g}^c_{e}$ and ${g}^p_{e}$ are gates of character and pinyin, respectively. $W^c_{g}$, $W^p_{g}$, $b^c_{g}$ and $b^p_{g}$ are model parameters. $\sigma$ denotes the sigmoid function. Then, we compute element-wise multiplication of character and pinyin representations with their corresponding gates:
\begin{equation}
\begin{split}
\label{eq:input_after_gate}
\bm{x}^c_{e}={W^c_{g}}\odot \bm{x}^c_e\\
\bm{x}^p_{e}={W^p_{g}}\odot \bm{x}^p_e
\end{split}
\end{equation}
Finally, we concatenate the $\bm{x}^c_{e}$ and $\bm{x}^p_{e}$ as final a fusion character representation.
\begin{equation}
\label{eq:fusion_input}
\bm{x}_{e}=[\bm{x}^c_{e},\bm{x}^p_{e}]
\end{equation}
\subsection{Lattice LSTM Layer}
\subsubsection{Baseline LSTM layer}
LSTM is an advanced recurrent neural network (RNN) with extra memory cells which are used to keep the long-term information and alleviate the gradient vanishing/exploding problem \cite{hochreiter_long_1997}. The basic LSTM functions are as follows:
\begin{equation}
\label{eq:lstm}
\begin{array}{c}
\left[ \begin{array}{l}
{\bf{o}}_e\\
{\bf{f}}_e\\
{\bf{\tilde c}}_e
\end{array} \right] = \left[ \begin{array}{c}
\sigma \\
\sigma \\
\tanh 
\end{array} \right]\left( {{{\bf{W}}^T}\left[ \begin{array}{l}
{\bf{x}}_e\\
{\bf{h}}_{e - 1}
\end{array} \right] + {\bf{b}}} \right)\\
{\bf{i}}_e = {\bf{1}} - {\bf{f}}_e \quad \quad \quad \quad\quad\\
{\bf{c}}_e = {\bf{f}}_e \odot {\bf{c}}_{e - 1} + {\bf{i}}_e \odot {\bf{\tilde c}}_e\\
{\bf{h}}_e{\rm{ = }}{\bf{o}}_e \odot \tanh ({\bf{c}}_e)\quad\quad
\end{array}
\end{equation}
where $h_e$ is the hidden vector of character $c_e$. $\bm{c}_e$ is the memory cell of character $c_e$. $\bm{i}_e$, $\bm{f}_e$ and $\bm{o}_e$ denote a set of the input, forget and output gates, respectively. Different from traditional LSTM, we set the input gate $i_e=1-f_e$,  according to \cite{greff_lstm:_2017}.
$\bf{W}$ and $\bf{b}$ are model parameters. $\sigma$ denotes the sigmoid function and $tanh$ denotes the hyperbolic tangent function.

\subsubsection{From LSTM  to Lattice LSTM }
\begin{figure}[h]
	\centering
	\includegraphics[width=0.60\linewidth]{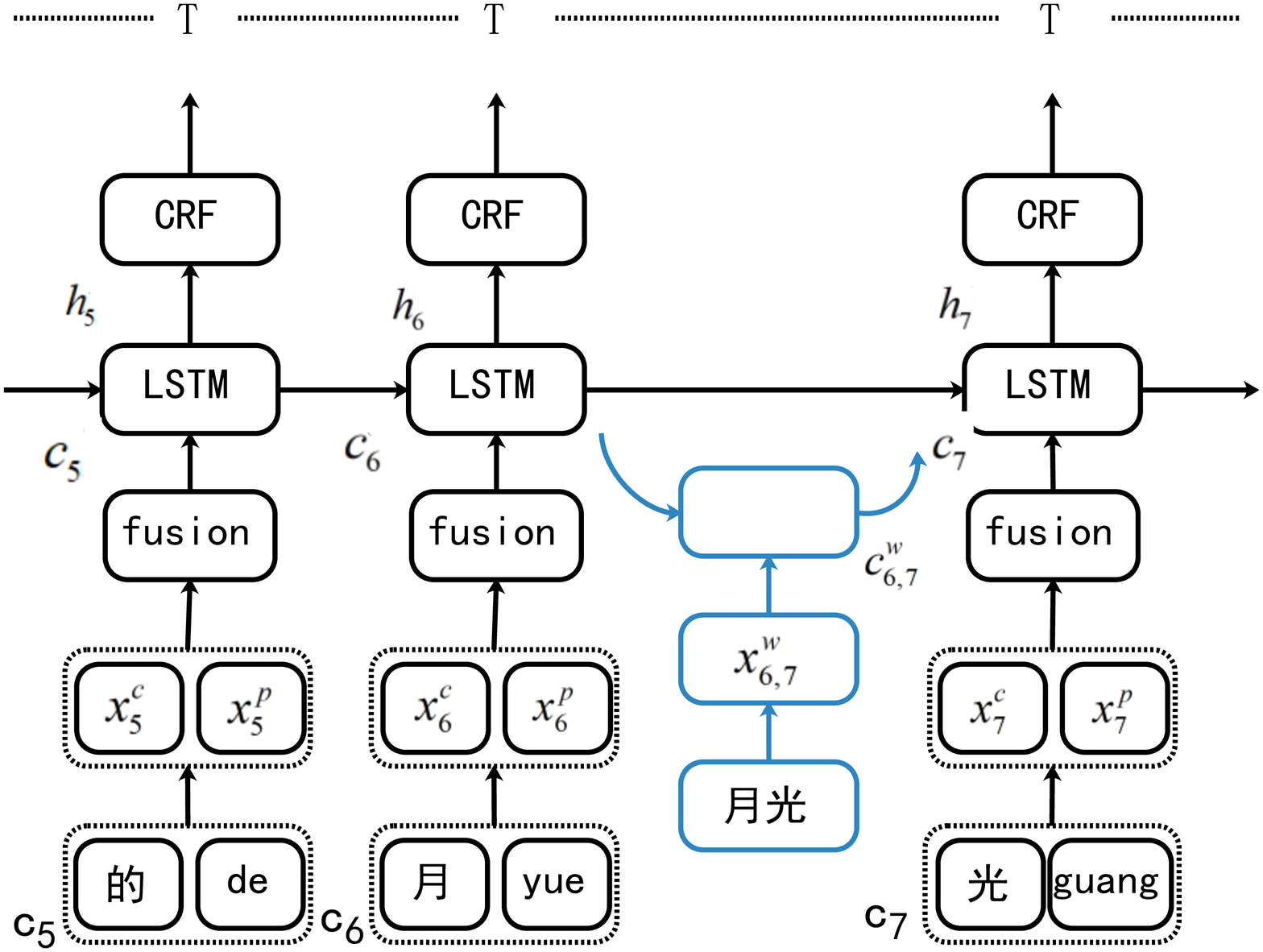}
	\caption{\label{fig:formula} Word Lattice}
\end{figure}

The lattice LSTM layer is an extension of the basic LSTM layer by adding “shortcut paths” which we name word lattices (blue parts in Figure \ref{fig:formula}). The inputs of the lattice LSTM  are a character sequence and all subsequences which are matched words in the word vocabulary $\mathbb{D}^w$. $\mathbb{D}^w$ are created following the paper \cite{zhang_chinese_2018} and illustrated in detail  in  the experiments section.

 The word lattice which is a variant LSTM cell without output gate takes $x^w_{b,e}$ and $h_b$ as inputs. $x^w_{b,e}$ is the word representation of character subsequence $(c_b,...,c_e)$  (Equation (\ref{eq:word_emb})). $h_b$ is the hidden vector of the first character of this subsequence.  The word lattice is calculated as follows:
\begin{equation}
\label{eq:word-lattice}
\begin{array}{c}
\left[ \begin{array}{l}
{\bm{i}}_{b,e}^w\\
{\bm{f}}_{b,e}^w\\
{\bm{\tilde c}}_{b,e}^w
\end{array} \right] = \left[ \begin{array}{c}
\sigma \\
\sigma \\
\tanh 
\end{array} \right]\left( {{{{\bm{W}}^{w}}^T}\left[ \begin{array}{l}
{\bm{x}}_{b,e}^w\\
{\bm{h}}_b
\end{array} \right] + {{\bm{b}}^w}} \right)\\
{\bm{c}}_{b,e}^w = {\bm{f}}_{b,e}^w \odot {\bm{c}}_b + {\bm{i}}_{b,e}^w \odot {\bm{\tilde c}}_{b,e}^w
\end{array}
\end{equation}

where ${\bm{c}}^w_{b,e}$ is the memory cell of word lattice starting from index $b$ of character sequence to index $e$. ${\bm{W}}^{w}$ and $b^w$ are the model parameters. There is no output gate. 

The memory cell ${\bm{c}}^w_{b,e}$ links to the end character  $\bm{c}_e$ as one of inputs to calculate the $\bm{h}_e$ of character $c_e$. Character $c_e$ may have multiple memory cell links. As shown in Figure \ref{fig:formula},  ${\bm{c}}^w_{6,7}$ the memory cell of the word lattice “月光”  links to “光” the seventh character of the input sequence. 
We define all the word lattices which link to character $c_e$ as a set $\mathbb{C}^w_e=\{\bm{c}_{b,e}|b\in \{b^{'}|(c_{b^{'}},...,c_e )\in \mathbb{D}^w\}\}$. We assign a unique input gate for each word lattice to control its  information contribution as follows:
\begin{equation}
\label{eq:lattice-contribution}
{\bm{i}}_{_{b,e}}^{w} = \sigma ({{\bm{W}}_g^{\rm{w}}}^T\left[ \begin{array}{l}
{\bm{x}}_e\\
{\bm{c}}_{b,e}^w
\end{array} \right] + {{\bm{b}}_g^{w}})
\end{equation}

where $\bm{i}_{_{b,e}}^{w}$ is the input gate of $c^w_{b,e}$. $\bm{W}_g^w$ and $b^w_g$ are  parameters for input gate of word lattice.



Until now, we have all input gates related to character $c_e$, namely one $i_e$ and  several ${\bm{i}}^w_{b^{'},e}$.
We normalize all these input gates to 1, and obtain one $\alpha_e$ for $i_e$ and a set of ${\alpha}^w_{b^{'},e}$ for ${\bm{i}}^w_{b^{'},e}$. (Here, $b^{'}$ can be any index before e, and satisfies $(c_{b^{'}:e})\in \mathbb{D}^w$). Taking $b^{'}= b$ for example:
\begin{equation}
\begin{array}{l}
{\bf{\alpha }}_e = \frac{{\exp (i_e)}}{{\exp (i_e) + \sum\limits_{b'} {\exp (i_{b',e}^w) } }}\\
{\bf{\alpha }}_{b,e}^w = \frac{{\exp (i_{b,e}^w)}}{{\exp (i_e) + \sum\limits_{b'} {\exp (i_{b',e}^w)} }}\\
\end{array}
\end{equation}
The final lattice LSTM representation $h_e$ of character $c_e$ is calculated as follows:  
\begin{equation}
\begin{array}{c}
\left[ \begin{array}{l}
{\bf{o}}_e\\
{\bf{f}}_e\\
{\bf{\tilde c}}_e
\end{array} \right] = \left[ \begin{array}{c}
\sigma \\
\sigma \\
\tanh 
\end{array} \right]\left( {{{\bf{W}}^T}\left[ \begin{array}{l}
{\bf{x}}_e\\
{\bf{h}}_{e - 1}
\end{array} \right] + {\bf{b}}} \right)\\
{\bf{i}}_e = {\bf{1}} - {\bf{f}}_e\\
{\bf{c}}_e = {\bf{\alpha }}_e \odot \bf{\tilde c}_e + \sum\limits_{{{\bf{c}}_{b'...e}} \in {\mathbb{D}^w}} {{\bf{\alpha }}_{b',e}^w \odot {\bf{c}}_{b',e}^w} \\
{\bf{h}}_e{\rm{ = }}{\bf{o}}_e \odot \tanh ({\bf{c}}_e)
\end{array}\
\label{eq:lattice_lstm_final}
\end{equation}
where $\bm{W}^T$ and $\bm{b}$ are the model parameters which 
are the same with the standard LSTM. Compared with Equation (\ref{eq:lstm}), 
Equation (\ref{eq:lattice_lstm_final}) has a more complex memory calculation step 
which integrates both the standard character LSTM memory ${{\bm{\tilde c}}}_e$ and all the memory cells of word lattices in ${\mathbb{C}}^w_e$. 


\subsubsection{Bidirectional Lattice LSTM} Above, we define one-directional FL-LSTM. For the bidirectional FL-LSTM, the input sequence $(x_1,x_2,...,x_m)$ are feed to a bidirectional model formalized as Equation (\ref{eq:lattice_lstm_final}) to obtain left-to-right ${\overrightarrow{\bm{h}}_1},
{\overrightarrow{\bm{h}}_2},
...,
{\overrightarrow{\bm{h}}_m}$ and right-to-left ${\overleftarrow{\bm{h}}_1},
{\overleftarrow{\bm{h}}_2},
...,
{\overleftarrow{\bm{h}}_m}$, respectively. In order to represent complete information, the hidden vector representation of each character in the sequence is concatenation of both sides of hidden vectors.

\begin{equation}
\label{eq:hidden-concat}
{{\bm{h}}_e} = [{\overrightarrow{\bm{h}}_e};{\overleftarrow{\bm{h}}_e}]
\end{equation}

A standard CRF layer is applied to the hidden vector sequence  ${\bm{h}_1},{\bm{h}_2},...,{\bm{h}_m}$ for label tagging.

\subsection{CRF Layer}
This paper uses a standard CRF layer \cite{Lafferty2001ConditionalRF} 
on  top of the hidden state sequence $h_1, h_2, ..., h_m$. 
The probability of every predict label sequence $y=l_1, l_2, ..., l_m$ is

\begin{equation}
\label{eq:crf-pro}
p(y|s) = \frac{{\exp (\sum\limits_{i = 1}^m {({W^{{y_i}}}} {h_i} + T({y_{i - 1}},{y_i})))}}{{\sum\limits_{y' \in Y(s)} {\exp (\sum\limits_{i = 1}^m {({W^{y'_i}}} {h_i} + T(y'_{i - 1},y'_i)))} }}
\end{equation}

where $Y(s)$ is the set of all possible label sequences on 
sentence s, and $y'$ is an arbitrary label sequence. $W^{y_i}$ denotes a 
model parameter specific to $l_i$, and $T(y_{i-1},y_i)$ represents the transition score from label $y_{i-1}$ to $y_i$.

\subsection{Decoding and Training}

Given a set of manually labeled training data $\{(s_i, y_i)\}$, sentence level 
log-likelihood loss is used to train the model:
\begin{equation}
\label{eq:crf-loss}
L(W,T) = \sum\limits_i {\log p({y_i}|{s_i};W,T)}
\end{equation}
To decode the highest scored label sequence over the input sentence, 
we adopt the Viterbi algorithm \cite{Viterbi1967ErrorBF}.  
Decoding is to search for the label sequence $y^*$ with the highest 
conditional probability:
\begin{equation}
\label{eq:crf-argmax}
{y^*} = \arg {\max _{y \in Y(s)}}p(y|s;W,T)
\end{equation}

\section{Experiments}
\label{sec:experiments}

We carry out a wide range of experiments on different datasets to investigate
the effectiveness of our model. What's more, we make an effort to compare the
baseline and our model under different settings. We use the standard precision (P), recall (R) and F1-score  as evaluation metrics.

\subsection{Experiment settings}
\label{subsec:settings}

\begin{table}[h]
\centering
\label{table:sighan_dataset}
\begin{tabular}{lllll}
\hline
\multicolumn{2}{l}{}                                                                        & Error \# & Char \#  & Sent \# \\ \hline
{2013}                       & \begin{tabular}[c]{@{}l@{}}train \scriptsize{13}\end{tabular} & 324     & 17,611  & 350    \\ \cline{2-5} 
                                            & \begin{tabular}[c]{@{}l@{}}test \scriptsize{13}\end{tabular} & 966     & 75,328  & 1,000  \\ \hline
{2014}                       & \begin{tabular}[c]{@{}l@{}}train \scriptsize{14}\end{tabular} & 5,224   & 330,656 & 6,527  \\ \cline{2-5} 
                                            & \begin{tabular}[c]{@{}l@{}}test \scriptsize{14}\end{tabular} & 510     & 54,176  & 1,063  \\ \hline
\multicolumn{1}{c}{{2015}} & \begin{tabular}[c]{@{}l@{}}train \scriptsize{15}\end{tabular} & 3,101   & 95,114  & 3,174  \\ \cline{2-5} 
\multicolumn{1}{c}{}                      & \begin{tabular}[c]{@{}l@{}}test \scriptsize{15}\end{tabular} & 531     & 34811   & 1,100  \\ \hline
\end{tabular}
\caption{Statistics of SIGHAN bakeoff datasets. Error
\# denotes the number of spelling errors, Char
\# represents the number of Chinese characters and Sent \# refers to the number of sentences.}
\end{table}

\subsubsection{Datasets} SIGHAN Bake-off datasets (2013-2015) are used as experimental data. 2013 is collected from the essays
written by native Chinese speakers \cite{Wu2013ChineseSC}. 2014 and 2015 are collected from essays written by learners of Chinese as a Foreign Language (CFL)   \cite{Yu2014ChineseSE,hsieh2015correcting}. Statistics of the datasets are shown in Table 1. These SIGHAN datasets are written in traditional Chinese, in order to match with the word lattice vocabulary extracted from a simplified Chinese embedding \cite{zhang_chinese_2018}, we convert these datasets from traditional Chinese to simplified Chinese using an online converter \footnote{ https://tool.lu/zhconvert/}. 

We split the SIGHAN data  based on the sentence ending symbols. The segmented sentences in train2013-2015 and  test2013-test2014 are merged as the train set, and test2015 is left as a test set. Finally, we obtain a train set with about 13k sentences and a test set with about 1k sentences.



\subsubsection{Vocabulary and Embedding} 

In this paper, three vocabularies are used, namely character vocabulary, word vocabulary and pinyin vocabulary. Character vocabulary is created when the training data is loaded. Pinyin vocabulary is obtained by converting  characters in character vocabulary to pinyin. Word vocabulary is derived from a pretrained word embedding mentioned in \cite{zhang_chinese_2018}, which contains 704.4k words among which the single word, the double words and multi-words are 5.7k, 291.5k and 278.1k, respectively. 

The weight of the corresponding embeddings of these three vocabularies  are assigned by random initialization.

\begin{table}[h]
\centering
\begin{tabular}{ll||ll}
\hline
\textbf{parameter} & \textbf{value} & \textbf{parameter} & \textbf{value} \\ \hline
char\_emb\_size    & 50             & word\_emb\_size    & 50             \\
pinyin\_emb\_size      & 50            & dropout    & 0.5            \\
LSTM layer         & 1              & LSTM hidden        & 200            \\
learning rate      & 0.015          & lr decay           & 0.05           \\ \hline
\end{tabular}
\caption{\label{tb:parameters} Hyper-parameters values. }
\end{table}
\subsubsection{Hyperparameters} Table \ref{tb:parameters} shows the values of hyperparameters of our models,  which  is  fixed  during  experiments without grid-search for each dataset \cite{zhang_chinese_2018}.
Adam optimization algorithm \cite{kingma2014adam} is adopted as the optimizer, with an initial rate of  0.015 and a decay rate of  0.05. The size of the above three embeddings is all
set to 50. The hidden size of LSTM is set to 200. Dropout
is applied to these three embeddings with a rate of 0.5.
One layer bi-direction LSTM is adopted.

\subsection{Result}

\begin{table}[h]
\centering
\setlength\arrayrulewidth{0.7pt}        
\begin{tabular}{|l|l|l|l|}
\hline
                                                                   & Pre.                                               & Rec.                                               & F1                                                          \\ \hline 
LSTM-CRF                                                           & 0.513                                              & 0.170                                              & 0.255                                                       \\ \hline
SCAU                                                               & 0.411                                              & 0.236                                              & 0.287                                                       \\ \hline
KUAS                                                               & \multicolumn{1}{r|}{0.538}                         & \multicolumn{1}{r|}{0.241}                         & \multicolumn{1}{r|}{0.332}                                  \\ \hline
NTOU                                                               & 0.543                                              & 0.299                                              & 0.352                                                       \\ \hline
NCTU\&NTUT                                                         & \multicolumn{1}{r|}{0.789}                         & \multicolumn{1}{r|}{0.294}                         & \multicolumn{1}{r|}{0.426}                                  \\ \hline
L-LSTM-CRF       & \multicolumn{1}{r|}{0.652} & \multicolumn{1}{r|}{0.364} & \multicolumn{1}{r|}{0.467}          \\ \hline
\textbf{FL-LSTM-CRF} & \multicolumn{1}{r|}{0.577} & \multicolumn{1}{r|}{0.427} & \multicolumn{1}{r|}{\textbf{0.491}} \\ \hline
\end{tabular}
\caption{Comparing test result on various models}
\label{tb:result}
\end{table}

We experimented on the SIGHAN dataset and compared our
models with four latest models mentioned in SIGHAN2015
bake-off competition. 

As shown in Table \ref{tb:result}, our best model,
FL-LSTM-CRF,  obtains  the  maximum  F1-score   of  0.491,
and  outperforms  all  other  models.  
However,  our  baseline model LSTM-CRF performs worst, and obtains the minimum F1-score  of 0.255. Our L-LSTM-CRF obtains the second maximum F1-score  of 0.467.

The above result confirms the availability of an end-to-end
framework  for  spelling  error  detection.  Only  adding  word
information by lattice, models based on end-to-end framework consistently outperform the pipeline models. Word lattices are very effective for error detection, which would improve F1-score  very obviously. Pinyin is  excellent supplement
information, and integrating pinyin information can further
improve model performance.

\subsection{Analysis}

\subsubsection{F1 against various error types} 
\begin{table}[h]
	\centering
	\begin{tabular}{|l|l|l|l|}
		\hline
		& homophone & near-homo & Others \\ \hline
		LSTM-CRF    & 0.279     & 0.174          & 0.279  \\ \hline
		L-LSTM-CRF  & 0.529     & 0.439          & 0.176  \\ \hline
		\textbf{FL-LSTM-CRF} & 0.571     & 0.484          & 0.196  \\ \hline
	\end{tabular}
	\caption{F1 against various error types}
	\label{tb:f1_against_types}
\end{table}

According to \cite{Liu2011VisuallyAP}, Chinese  errors  are  roughly  divided  into  two  types: phonological  errors  and  visual  errors.  The phonological errors  can  be  future  divided  into  homophone  errors  and near-homophone  errors.  The  homophone  errors  are  characters  which have  some  pinyin  but  different  tones,  for  example, 交 (jiao, tone 1)  and 叫 (jiao, tone 4). Near-homophone errors are characters which have the similar pinyin, for example, 是 (shi) and 思 (si). Visual errors refer to errors having similar shapes, such as “太(tai)” and “大(da)”. 

As shown in Table \ref{tb:f1_against_types}, experiments on different error types show   after adding pinyin and word information, the model achieves significant improvement in homophone and near-homophone errors, which proves that word and pinyin information is availability  for error detection \cite{Yang2012SpellCF}. However, adding pinyin has no obvious effect for other type of errors, because phonological errors are caused by the use of the pinyin input method. It has a close relationship with pinyin, so adding pinyin information can significantly improve the performance. However, other errors, such as visual errors, are caused by OCR recognition and have little relation with pinyin.



\subsubsection{Case Study} 

\begin{table}[h]
\centering
\begin{tabular}{|l|l|l|l|l|}
\hline
                                                         & \multicolumn{2}{c|}{\textbf{Example 1}}                                                                                   & \multicolumn{2}{c|}{\textbf{Example 2}}                                                                                                           \\ \hline
\begin{tabular}[c]{@{}l@{}}correct\\ source\end{tabular} & \multicolumn{2}{l|}{\begin{tabular}[c]{@{}l@{}}漂/亮/的/女/孩/子\\ piao/liang/de/nv/hai/zi\\ beautiful girl\end{tabular}}       & \multicolumn{2}{l|}{\begin{tabular}[c]{@{}l@{}}在/那/里/野/餐/也/不/错\\ zai/na/li/ye/can/ye/bu/cuo\\ It’s also good to have a picnic there\end{tabular}} \\ \hline
\begin{tabular}[c]{@{}l@{}}input\\ source\end{tabular}   & \multicolumn{2}{l|}{\begin{tabular}[c]{@{}l@{}}漂/亮/的/\textcolor{red}{努}/孩/子\\ piao/liang/de/\textcolor{red}{nu}/hai/zi\\ beautiful /\textcolor{red}{hard} /child\end{tabular}} & \multicolumn{2}{l|}{\begin{tabular}[c]{@{}l@{}}在/\textcolor{red}{哪}/里/野/餐/也/不/错\\ zai/\textcolor{red}{na}/li/ye/can/ye/bu/cuo\\ in / \textcolor{red}{where} / picnic / also / nice\end{tabular}}            \\ \hline
\multicolumn{5}{|c|}{Predict Result}                                                                                                                                                                                                                                                                                                     \\ \hline
LSTM-CRF                                                 & \begin{tabular}[c]{@{}l@{}}漂/亮/的/努/孩/子\\ piao/liang/de/nu/hai/zi\\ beautiful/hard/child\end{tabular}          & ×         & \begin{tabular}[c]{@{}l@{}}在/哪/里/野/餐/也/不/错\\ zai/na/li/ye/can/ye/bu/cuo\\ in/where/picnic/also/nice\end{tabular}                & ×               \\ \hline
L-LSTM-CRF                                               &  \begin{tabular}[c]{@{}l@{}}\colorbox{yellow}{漂/亮}/的  /\textcolor{green}{努}/ \colorbox{yellow}{孩/子} \\ \colorbox{yellow}{piao/liang}/de  /\textcolor{green}{nu}/ \colorbox{yellow}{hai/zi}\\ \colorbox{yellow}{beautiful}/\textcolor{green}{hard}/ \colorbox{yellow}{child} \end{tabular}         & √         & \begin{tabular}[c]{@{}l@{}}在/\colorbox{yellow}{哪/里}/\colorbox{yellow}{野/餐}/也/\colorbox{yellow}{不/错}\\ zai/\colorbox{yellow}{na/li}/\colorbox{yellow}{ye/can}/ye/\colorbox{yellow}{bu/cuo}\\ in/\colorbox{yellow}{where}/\colorbox{yellow}{picnic}/also/\colorbox{yellow}{nice}  \end{tabular}                & ×               \\ \hline
\textbf{FL-LSTM-CRF}                                     & \begin{tabular}[c]{@{}l@{}}\colorbox{yellow}{漂/亮}/的  /\textcolor{green}{努}/ \colorbox{yellow}{孩/子} \\ \colorbox{yellow}{piao/liang}/de  /\textcolor{green}{nu}/ \colorbox{yellow}{hai/zi} \\ \colorbox{yellow}{beautiful} /\textcolor{green}{hard}/ \colorbox{yellow}{child} \end{tabular}          & √         & \begin{tabular}[c]{@{}l@{}} 在/\colorbox{yellow}{\textcolor{green}{哪}/里}/\colorbox{yellow}{野/餐}/也/\colorbox{yellow}{不/错} \\ zai/\colorbox{yellow}{\textcolor{green}{na}/li}/\colorbox{yellow}{ye/can}/ye/\colorbox{yellow}{bu/cuo} \\ in/\colorbox{yellow}{\textcolor{green}{where}}/\colorbox{yellow}{picnic}/also/\colorbox{yellow}{nice}  \end{tabular}                & √               \\ \hline
\end{tabular}
\caption{\label{tb:case-study} Case study.  Red color represents real misspelling characters. Green represents misspelling characters predicted by models. Yellow color represents the matched word lattices}
\end{table}

Table \ref{tb:case-study} shows two examples of error detection using various models: 

In example 1, we detect errors from the sentence “漂亮的努孩子(beautiful girl)” with a spelling error “努(nu)” using various models. LSTM-CRF model fails to detect it, while L-LSTM-CRF and FL-LSTM-CRF both detect it successfully, which proves the advantages of word lattices. The reason is that matched word lattices (marked with yellow), such as “漂亮(piao liang)” and “孩子(hai zi)” give extra information to models, and improve the probability of error detection by pointing out the correct words.

In example 2, we detect errors from sentence “在哪里野餐也不错(It’s also good to have a picnic there)” with a  spelling error “哪(na)”.   LSTM-CRF and L-LSTM-CRF both fail to detect the misspelled character “哪(na)”, while FL-LSTM-CRF catches this error. 
The misspelling character “哪(na)”  and  its corresponding right character “那(na)” share the same pinyin “na”, so it is a homophone error. 
For the L-LSTM-CRF model, the match word  lattice “哪里(na li)” provides a cue that  “哪(na)” may be a correct character, which leads to failure of error detecting. However,  the FL-LSTM-CRF detect this error correctly, although the matched word lattices also provide the misleading cue about “哪(na)”.  Character-pinyin-fusion inputs help FL-LSTM-CRF to  catch this error  successfully,  and pinyin information provides an extra cue that “哪(na)” in this context may be a misspelling of “那(na)”.

\section{Conclusion}
\label{sec:conclusion}
In this paper, we propose a novel spelling error detection model, called FL-LSTM-CRF, which is under the end-to-end framework without dividing the error detection process into steps. What is more, this model has the ability to integrate character, word and pinyin information together. Experiments on SIGHAN dataset  prove the applicability of the end-to-end framework, as well as the positive effect of fusion character, word and pinyin information for error detection.  FL-LSTM-CRF consistently outperform other models with  similar external resources. 

 In the future, we will study further on how to adopt more information to improve the performance of error detection and how to generalize our model to more languages.


\end{CJK*}
\bibliography{lattice-csc}


\begin{thebibliography}{27}


\ifx \showCODEN    \undefined \def \showCODEN     #1{\unskip}     \fi
\ifx \showDOI      \undefined \def \showDOI       #1{#1}\fi
\ifx \showISBNx    \undefined \def \showISBNx     #1{\unskip}     \fi
\ifx \showISBNxiii \undefined \def \showISBNxiii  #1{\unskip}     \fi
\ifx \showISSN     \undefined \def \showISSN      #1{\unskip}     \fi
\ifx \showLCCN     \undefined \def \showLCCN      #1{\unskip}     \fi
\ifx \shownote     \undefined \def \shownote      #1{#1}          \fi
\ifx \showarticletitle \undefined \def \showarticletitle #1{#1}   \fi
\ifx \showURL      \undefined \def \showURL       {\relax}        \fi
\providecommand\bibfield[2]{#2}
\providecommand\bibinfo[2]{#2}
\providecommand\natexlab[1]{#1}
\providecommand\showeprint[2][]{arXiv:#2}

\bibitem[\protect\citeauthoryear{Burstein and Chodorow}{Burstein and
  Chodorow}{1999}]%
        {Burstein1999AutomatedES}
\bibfield{author}{\bibinfo{person}{Jill Burstein} {and} \bibinfo{person}{Martin
  Chodorow}.} \bibinfo{year}{1999}\natexlab{}.
\newblock \showarticletitle{Automated Essay Scoring For Nonnative English
  Speakers}.
\newblock


\bibitem[\protect\citeauthoryear{Chang}{Chang}{1995}]%
        {chang1995new}
\bibfield{author}{\bibinfo{person}{Chao-Huang Chang}.}
  \bibinfo{year}{1995}\natexlab{}.
\newblock \showarticletitle{A new approach for automatic Chinese spelling
  correction}. In \bibinfo{booktitle}{\emph{Proceedings of Natural Language
  Processing Pacific Rim Symposium}}, Vol.~\bibinfo{volume}{95}. Citeseer,
  \bibinfo{pages}{278--283}.
\newblock


\bibitem[\protect\citeauthoryear{Chen, Lee, Lee, Wang, and Chen}{Chen
  et~al\mbox{.}}{2013}]%
        {chen2013study}
\bibfield{author}{\bibinfo{person}{Kuan-Yu Chen}, \bibinfo{person}{Hung-Shin
  Lee}, \bibinfo{person}{Chung-Han Lee}, \bibinfo{person}{Hsin-Min Wang}, {and}
  \bibinfo{person}{Hsin-Hsi Chen}.} \bibinfo{year}{2013}\natexlab{}.
\newblock \showarticletitle{A study of language modeling for Chinese spelling
  check}. In \bibinfo{booktitle}{\emph{Proceedings of the Seventh SIGHAN
  Workshop on Chinese Language Processing}}. \bibinfo{pages}{79--83}.
\newblock


\bibitem[\protect\citeauthoryear{Chiu, Wu, and Chang}{Chiu
  et~al\mbox{.}}{2013}]%
        {chiu2013chinese}
\bibfield{author}{\bibinfo{person}{Hsun-wen Chiu}, \bibinfo{person}{Jian-cheng
  Wu}, {and} \bibinfo{person}{Jason~S Chang}.} \bibinfo{year}{2013}\natexlab{}.
\newblock \showarticletitle{Chinese spelling checker based on statistical
  machine translation}. In \bibinfo{booktitle}{\emph{Proceedings of the Seventh
  SIGHAN Workshop on Chinese Language Processing}}. \bibinfo{pages}{49--53}.
\newblock


\bibitem[\protect\citeauthoryear{Gao, Li, Micol, Quirk, and Sun}{Gao
  et~al\mbox{.}}{2010}]%
        {Gao2010ALS}
\bibfield{author}{\bibinfo{person}{Jianfeng Gao}, \bibinfo{person}{Xiaolong
  Li}, \bibinfo{person}{Daniel Micol}, \bibinfo{person}{Chris Quirk}, {and}
  \bibinfo{person}{Xu Sun}.} \bibinfo{year}{2010}\natexlab{}.
\newblock \showarticletitle{A Large Scale Ranker-Based System for Search Query
  Spelling Correction}. In \bibinfo{booktitle}{\emph{COLING}}.
\newblock


\bibitem[\protect\citeauthoryear{Greff, Srivastava, Koutník, Steunebrink, and
  Schmidhuber}{Greff et~al\mbox{.}}{2017}]%
        {greff_lstm:_2017}
\bibfield{author}{\bibinfo{person}{Klaus Greff}, \bibinfo{person}{Rupesh~K.
  Srivastava}, \bibinfo{person}{Jan Koutník}, \bibinfo{person}{Bas~R.
  Steunebrink}, {and} \bibinfo{person}{Jürgen Schmidhuber}.}
  \bibinfo{year}{2017}\natexlab{}.
\newblock \showarticletitle{{LSTM}: {A} search space odyssey}.
\newblock \bibinfo{journal}{\emph{IEEE transactions on neural networks and
  learning systems}} \bibinfo{volume}{28}, \bibinfo{number}{10}
  (\bibinfo{year}{2017}), \bibinfo{pages}{2222--2232}.
\newblock


\bibitem[\protect\citeauthoryear{Han and Chang}{Han and Chang}{2013}]%
        {han2013maximum}
\bibfield{author}{\bibinfo{person}{Dongxu Han} {and} \bibinfo{person}{Baobao
  Chang}.} \bibinfo{year}{2013}\natexlab{}.
\newblock \showarticletitle{A maximum entropy approach to Chinese spelling
  check}. In \bibinfo{booktitle}{\emph{Proceedings of the Seventh SIGHAN
  Workshop on Chinese Language Processing}}. \bibinfo{pages}{74--78}.
\newblock


\bibitem[\protect\citeauthoryear{He and Fu}{He and Fu}{2013}]%
        {he2013description}
\bibfield{author}{\bibinfo{person}{Yu He} {and} \bibinfo{person}{Guohong Fu}.}
  \bibinfo{year}{2013}\natexlab{}.
\newblock \showarticletitle{Description of {HLJU} {C}hinese Spelling Checker
  for {SIGHAN} Bakeoff 2013}. In \bibinfo{booktitle}{\emph{SIGHAN@IJCNLP}}.
\newblock


\bibitem[\protect\citeauthoryear{Hochreiter and Schmidhuber}{Hochreiter and
  Schmidhuber}{1997}]%
        {hochreiter_long_1997}
\bibfield{author}{\bibinfo{person}{Sepp Hochreiter} {and}
  \bibinfo{person}{Jürgen Schmidhuber}.} \bibinfo{year}{1997}\natexlab{}.
\newblock \showarticletitle{Long short-term memory}.
\newblock \bibinfo{journal}{\emph{Neural computation}} \bibinfo{volume}{9},
  \bibinfo{number}{8} (\bibinfo{year}{1997}), \bibinfo{pages}{1735--1780}.
\newblock


\bibitem[\protect\citeauthoryear{Hsieh, Bai, and Chen}{Hsieh
  et~al\mbox{.}}{2013}]%
        {hsieh2013introduction}
\bibfield{author}{\bibinfo{person}{Yu-Ming Hsieh}, \bibinfo{person}{Ming-Hong
  Bai}, {and} \bibinfo{person}{Keh-Jiann Chen}.}
  \bibinfo{year}{2013}\natexlab{}.
\newblock \showarticletitle{Introduction to CKIP Chinese spelling check system
  for SIGHAN Bakeoff 2013 evaluation}. In \bibinfo{booktitle}{\emph{Proceedings
  of the Seventh SIGHAN Workshop on Chinese Language Processing}}.
  \bibinfo{pages}{59--63}.
\newblock


\bibitem[\protect\citeauthoryear{Hsieh, Bai, Huang, and Chen}{Hsieh
  et~al\mbox{.}}{2015}]%
        {hsieh2015correcting}
\bibfield{author}{\bibinfo{person}{Yu-Ming Hsieh}, \bibinfo{person}{Ming-Hong
  Bai}, \bibinfo{person}{Shu-Ling Huang}, {and} \bibinfo{person}{Keh-Jiann
  Chen}.} \bibinfo{year}{2015}\natexlab{}.
\newblock \showarticletitle{Correcting Chinese spelling errors with word
  lattice decoding}.
\newblock \bibinfo{journal}{\emph{ACM Transactions on Asian and Low-Resource
  Language Information Processing (TALLIP)}} \bibinfo{volume}{14},
  \bibinfo{number}{4} (\bibinfo{year}{2015}), \bibinfo{pages}{18}.
\newblock


\bibitem[\protect\citeauthoryear{Jie~Yang and Liang}{Jie~Yang and
  Liang}{2019}]%
        {yang_subword_2018}
\bibfield{author}{\bibinfo{person}{Yue~Zhang Jie~Yang} {and}
  \bibinfo{person}{Shuailong Liang}.} \bibinfo{year}{2019}\natexlab{}.
\newblock \showarticletitle{Subword Encoding in Lattice LSTM for Chinese Word
  Segmentation}. In \bibinfo{booktitle}{\emph{NAACL}}.
\newblock


\bibitem[\protect\citeauthoryear{Kingma and Ba}{Kingma and Ba}{2014}]%
        {kingma2014adam}
\bibfield{author}{\bibinfo{person}{Diederik~P Kingma} {and}
  \bibinfo{person}{Jimmy Ba}.} \bibinfo{year}{2014}\natexlab{}.
\newblock \showarticletitle{Adam: A method for stochastic optimization}.
\newblock \bibinfo{journal}{\emph{arXiv preprint arXiv:1412.6980}}
  (\bibinfo{year}{2014}).
\newblock


\bibitem[\protect\citeauthoryear{Lafferty, McCallum, and Pereira}{Lafferty
  et~al\mbox{.}}{2001}]%
        {Lafferty2001ConditionalRF}
\bibfield{author}{\bibinfo{person}{John~D. Lafferty}, \bibinfo{person}{Andrew
  McCallum}, {and} \bibinfo{person}{Fernando Pereira}.}
  \bibinfo{year}{2001}\natexlab{}.
\newblock \showarticletitle{Conditional Random Fields: Probabilistic Models for
  Segmenting and Labeling Sequence Data}. In \bibinfo{booktitle}{\emph{ICML}}.
\newblock


\bibitem[\protect\citeauthoryear{Liu, Lai, Tien, Chuang, Wu, and Lee}{Liu
  et~al\mbox{.}}{2011}]%
        {Liu2011VisuallyAP}
\bibfield{author}{\bibinfo{person}{C.-L. Liu}, \bibinfo{person}{M.-H. Lai},
  \bibinfo{person}{K.-W. Tien}, \bibinfo{person}{Y.-H. Chuang},
  \bibinfo{person}{S.-H. Wu}, {and} \bibinfo{person}{C.-Y. Lee}.}
  \bibinfo{year}{2011}\natexlab{}.
\newblock \showarticletitle{Visually and {Phonologically} {Similar}
  {Characters} in {Incorrect} {Chinese} {Words}: {Analyses}, {Identification},
  and {Applications}}.
\newblock \bibinfo{journal}{\emph{ACM Trans. Asian Lang. Inf. Process.}}
  \bibinfo{volume}{10} (\bibinfo{year}{2011}), \bibinfo{pages}{10:1--10:39}.
\newblock


\bibitem[\protect\citeauthoryear{Liu, Cheng, Luo, Duh, and Matsumoto}{Liu
  et~al\mbox{.}}{2013}]%
        {liu2013hybrid}
\bibfield{author}{\bibinfo{person}{Xiaodong Liu}, \bibinfo{person}{Kevin
  Cheng}, \bibinfo{person}{Yanyan Luo}, \bibinfo{person}{Kevin Duh}, {and}
  \bibinfo{person}{Yuji Matsumoto}.} \bibinfo{year}{2013}\natexlab{}.
\newblock \showarticletitle{A hybrid Chinese spelling correction using language
  model and statistical machine translation with reranking}. In
  \bibinfo{booktitle}{\emph{Proceedings of the Seventh SIGHAN Workshop on
  Chinese Language Processing}}. \bibinfo{pages}{54--58}.
\newblock


\bibitem[\protect\citeauthoryear{Lonsdale and Strong-Krause}{Lonsdale and
  Strong-Krause}{2003}]%
        {Lonsdale2003AutomatedRO}
\bibfield{author}{\bibinfo{person}{Deryle~W. Lonsdale} {and}
  \bibinfo{person}{Diane Strong-Krause}.} \bibinfo{year}{2003}\natexlab{}.
\newblock \showarticletitle{Automated Rating Of ESL Essays}. In
  \bibinfo{booktitle}{\emph{HLT-NAACL 2003}}.
\newblock


\bibitem[\protect\citeauthoryear{Ma and Hovy}{Ma and Hovy}{2016}]%
        {ma_end--end_2016}
\bibfield{author}{\bibinfo{person}{Xuezhe Ma} {and} \bibinfo{person}{Eduard
  Hovy}.} \bibinfo{year}{2016}\natexlab{}.
\newblock \showarticletitle{End-to-end {Sequence} {Labeling} via
  {Bi}-directional {LSTM}-{CNNs}-{CRF}}. In
  \bibinfo{booktitle}{\emph{Proceedings of the 54th {Annual} {Meeting} of the
  {Association} for {Computational} {Linguistics} ({Volume} 1: {Long}
  {Papers})}}. \bibinfo{publisher}{Association for Computational Linguistics},
  \bibinfo{address}{Berlin, Germany}, \bibinfo{pages}{1064--1074}.
\newblock
\urldef\tempurl%
\url{https://doi.org/10.18653/v1/P16-1101}
\showDOI{\tempurl}


\bibitem[\protect\citeauthoryear{Martins and Silva}{Martins and Silva}{2004}]%
        {Martins2004SpellingCF}
\bibfield{author}{\bibinfo{person}{Bruno Martins} {and}
  \bibinfo{person}{M{\'a}rio~J. Silva}.} \bibinfo{year}{2004}\natexlab{}.
\newblock \showarticletitle{Spelling Correction for Search Engine Queries}. In
  \bibinfo{booktitle}{\emph{EsTAL}}.
\newblock


\bibitem[\protect\citeauthoryear{Viterbi}{Viterbi}{1967}]%
        {Viterbi1967ErrorBF}
\bibfield{author}{\bibinfo{person}{Andrew~J. Viterbi}.}
  \bibinfo{year}{1967}\natexlab{}.
\newblock \showarticletitle{Error bounds for convolutional codes and an
  asymptotically optimum decoding algorithm}.
\newblock \bibinfo{journal}{\emph{IEEE Trans. Information Theory}}
  \bibinfo{volume}{13} (\bibinfo{year}{1967}), \bibinfo{pages}{260--269}.
\newblock


\bibitem[\protect\citeauthoryear{Wang, Song, Li, Han, and Zhang}{Wang
  et~al\mbox{.}}{2018}]%
        {wang_hybrid_2018}
\bibfield{author}{\bibinfo{person}{Dingmin Wang}, \bibinfo{person}{Yan Song},
  \bibinfo{person}{Jing Li}, \bibinfo{person}{Jialong Han}, {and}
  \bibinfo{person}{Haisong Zhang}.} \bibinfo{year}{2018}\natexlab{}.
\newblock \showarticletitle{A {Hybrid} {Approach} to {Automatic} {Corpus}
  {Generation} for {Chinese} {Spelling} {Check}}. In
  \bibinfo{booktitle}{\emph{Proceedings of the 2018 {Conference} on {Empirical}
  {Methods} in {Natural} {Language} {Processing}, {Brussels}, {Belgium},
  {October} 31 - {November} 4, 2018}}, \bibfield{editor}{\bibinfo{person}{Ellen
  Riloff}, \bibinfo{person}{David Chiang}, \bibinfo{person}{Julia Hockenmaier},
  {and} \bibinfo{person}{Jun'ichi Tsujii}} (Eds.).
  \bibinfo{publisher}{Association for Computational Linguistics},
  \bibinfo{pages}{2517--2527}.
\newblock
\showISBNx{978-1-948087-84-1}
\urldef\tempurl%
\url{https://aclanthology.info/papers/D18-1273/d18-1273}
\showURL{%
\tempurl}


\bibitem[\protect\citeauthoryear{Wu, Liu, and Lee}{Wu et~al\mbox{.}}{2013}]%
        {Wu2013ChineseSC}
\bibfield{author}{\bibinfo{person}{Shih-Hung Wu}, \bibinfo{person}{Chao-Lin
  Liu}, {and} \bibinfo{person}{Lung-Hao Lee}.} \bibinfo{year}{2013}\natexlab{}.
\newblock \showarticletitle{Chinese Spelling Check Evaluation at SIGHAN
  Bake-off 2013}. In \bibinfo{booktitle}{\emph{SIGHAN@IJCNLP}}.
\newblock


\bibitem[\protect\citeauthoryear{Yang, Zhao, Wang, and Lu}{Yang
  et~al\mbox{.}}{2012}]%
        {Yang2012SpellCF}
\bibfield{author}{\bibinfo{person}{Shaohua Yang}, \bibinfo{person}{Hai Zhao},
  \bibinfo{person}{Xiaolin Wang}, {and} \bibinfo{person}{Bao-Liang Lu}.}
  \bibinfo{year}{2012}\natexlab{}.
\newblock \showarticletitle{Spell Checking for Chinese}. In
  \bibinfo{booktitle}{\emph{LREC}}.
\newblock


\bibitem[\protect\citeauthoryear{Yang, Hsieh, Chen, Tsang, Shih, and Hsu}{Yang
  et~al\mbox{.}}{2013}]%
        {yang2013sinica}
\bibfield{author}{\bibinfo{person}{Ting-Hao Yang}, \bibinfo{person}{Yu-Lun
  Hsieh}, \bibinfo{person}{Yu-Hsuan Chen}, \bibinfo{person}{Michael Tsang},
  \bibinfo{person}{Cheng-Wei Shih}, {and} \bibinfo{person}{Wen-Lian Hsu}.}
  \bibinfo{year}{2013}\natexlab{}.
\newblock \showarticletitle{Sinica-IASL Chinese spelling check system at
  SIGHAN-7}. In \bibinfo{booktitle}{\emph{Proceedings of the Seventh SIGHAN
  Workshop on Chinese Language Processing}}. \bibinfo{pages}{93--96}.
\newblock


\bibitem[\protect\citeauthoryear{Yeh, Li, Wu, Chen, and Su}{Yeh
  et~al\mbox{.}}{2013}]%
        {yeh2013chinese}
\bibfield{author}{\bibinfo{person}{Jui-Feng Yeh}, \bibinfo{person}{Sheng-Feng
  Li}, \bibinfo{person}{Mei-Rong Wu}, \bibinfo{person}{Wen-Yi Chen}, {and}
  \bibinfo{person}{Mao-Chuan Su}.} \bibinfo{year}{2013}\natexlab{}.
\newblock \showarticletitle{Chinese word spelling correction based on n-gram
  ranked inverted index list}. In \bibinfo{booktitle}{\emph{Proceedings of the
  Seventh SIGHAN Workshop on Chinese Language Processing}}.
  \bibinfo{pages}{43--48}.
\newblock


\bibitem[\protect\citeauthoryear{Yu and Li}{Yu and Li}{2014}]%
        {Yu2014ChineseSE}
\bibfield{author}{\bibinfo{person}{Junjie Yu} {and} \bibinfo{person}{Zhenghua
  Li}.} \bibinfo{year}{2014}\natexlab{}.
\newblock \showarticletitle{Chinese Spelling Error Detection and Correction
  Based on Language Model, Pronunciation, and Shape}. In
  \bibinfo{booktitle}{\emph{CIPS-SIGHAN}}.
\newblock


\bibitem[\protect\citeauthoryear{Zhang and Yang}{Zhang and Yang}{2018}]%
        {zhang_chinese_2018}
\bibfield{author}{\bibinfo{person}{Yue Zhang} {and} \bibinfo{person}{Jie
  Yang}.} \bibinfo{year}{2018}\natexlab{}.
\newblock \showarticletitle{Chinese {NER} {Using} {Lattice} {LSTM}}. In
  \bibinfo{booktitle}{\emph{ACL}}.
\newblock


\end{thebibliography}
\bibliographystyle{ACM-Reference-Format}
\end{document}